\documentclass{article}
\usepackage{amsmath,amssymb,amsthm}  % gives theorem/proof environments
\usepackage{fontawesome5}
\usepackage{tikz}
\usepackage[mathlines]{lineno}
\usepackage{threeparttable}

\usetikzlibrary{positioning,arrows.meta,fit,calc,shadows.blur}
\DeclareUnicodeCharacter{2014}{, } %long dash converter
\definecolor{nuPurple}{HTML}{4E2A84}
\definecolor{nuLight}{HTML}{EEF0F7}
\definecolor{panel}{HTML}{FAFBFD}
\definecolor{teal}{HTML}{00A7A7}
\definecolor{pink}{HTML}{E94E77}
\definecolor{indigo}{HTML}{3B82F6}
\definecolor{okgreen}{HTML}{5AA469}
\definecolor{amber}{HTML}{F59E0B}
\definecolor{softline}{RGB}{210,214,220}
\usepackage{amsmath}
\usepackage{amssymb}
\usepackage{placeins}
\usepackage{wrapfig}
\usepackage{wrapfig}
\usepackage{enumitem}

\theoremstyle{plain}  % bold title, italic body

\usepackage{placeins}
\theoremstyle{definition} % bold title, upright body

\theoremstyle{remark} % italic title, upright body
\usepackage{float}

\usepackage{appendix} % Include the appendix package
\usepackage{float}
\usepackage{PRIMEarxiv}
\usepackage{algorithm}
\usepackage{algpseudocode}
\usepackage[utf8]{inputenc} % allow utf-8 input
\usepackage{bbm}

\usepackage[T1]{fontenc}    % use 8-bit T1 fonts
\usepackage{hyperref}       % hyperlinks
\usepackage{url}            % simple URL typesetting
\usepackage{booktabs}       % professional-quality tables
\usepackage{amsfonts}       % blackboard math symbols
\usepackage{nicefrac}       % compact symbols for 1/2, etc.
\usepackage{microtype}      % microtypography
\usepackage{lipsum}
\usepackage{fancyhdr}       % header
\usepackage{graphicx}       % graphics
\graphicspath{{media/}}    
\usepackage{color,soul}
\usepackage{xcolor}
\usepackage{amsmath}
\usepackage{longtable}
\usepackage{tabularx}
\usepackage{array}
\usepackage{tabularx}
\usepackage{array}
\usepackage{booktabs}
\newcolumntype{Y}{>{\raggedright\arraybackslash}X}
\newcolumntype{L}[1]{>{\raggedright\arraybackslash}p{#1}}

\title{ConvFormer3D-TAP: Phase/Uncertainty-Aware Front-End Fusion for Cine CMR View Classification Pipelines}
\author{%
\parbox{\textwidth}{\centering
Nafiseh~Ghaffar~Nia\textsuperscript{1,2,3,4},
Vinesh~Appadurai\textsuperscript{5},
Suchithra~V.\textsuperscript{6},
Chinmay~Rane\textsuperscript{6},
Daniel~Pittman\textsuperscript{1,2,3,4},
James~Carr\textsuperscript{1,2,3,4},
and~Adrienne~Kline\textsuperscript{1,2,3,4,*} \\
\textsuperscript{1}Center for Artificial Intelligence, Bluhm Cardiovascular Institute, Northwestern Medicine, Chicago, IL, USA \\
\textsuperscript{2}Department of Electrical and Computer Engineering, Northwestern University, Chicago, IL, USA \\
\textsuperscript{3}Department of Surgery, Northwestern University, Chicago, IL, USA \\
\textsuperscript{4}Department of Radiology, Northwestern University, Chicago, IL, USA; and Xtasis Inc., Chicago, IL, USA \\
\textsuperscript{5}Department of Cardiology, The Prince Charles Hospital; and School of Medicine, The University of Queensland, Queensland, Australia \\
\textsuperscript{6}Quantiphi Inc., USA \\
\textsuperscript{*}Corresponding author: \texttt{adrienne.kline@northwestern.edu}
}%
}

\begin{document}
\maketitle
\begin{abstract}
Reliable recognition of standard cine cardiac MRI views is essential because each view determines which cardiac anatomy is visualized and which quantitative analyses can be performed. Incorrect view identification, whether by a human reader or an automated deep learning system, can propagate errors into segmentation, volumetric assessment, strain analysis, and valve evaluation. However, accurate view classification remains challenging under routine clinical variability in scanner vendor, acquisition protocol, motion artifacts, and plane prescription. We present ConvFormer3D-TAP, a cine-specific spatiotemporal architecture that integrates 3D convolutional tokenization with multiscale self-attention. The model is trained using masked spatiotemporal reconstruction and uncertainty-weighted multi-clip fusion to enhance robustness across cardiac phases and ambiguous temporal segments. The design captures complementary cues: local anatomical structure through convolutional priors and long-range cardiac-cycle dynamics through hierarchical attention.
On a cohort of 150,974 clinically acquired cine sequences spanning six standard cine cardiac MRI views, ConvFormer3D-TAP achieved 96\% validation accuracy with per-class F1-scores $\geq 0.94$ and strong calibration (ECE = 0.025; Brier = 0.040). Error analysis shows that residual confusions are concentrated in anatomically adjacent long-axis and LVOT/AV view pairs, consistent with intrinsic prescription overlap. These results support ConvFormer3D-TAP as a scalable front-end for view routing, filtering and quality control in end-to-end cMRI workflows.
\end{abstract}

% keywords can be removed

\section{Introduction}

\begin{figure}[t]
\centering
\includegraphics[width=0.4\columnwidth]{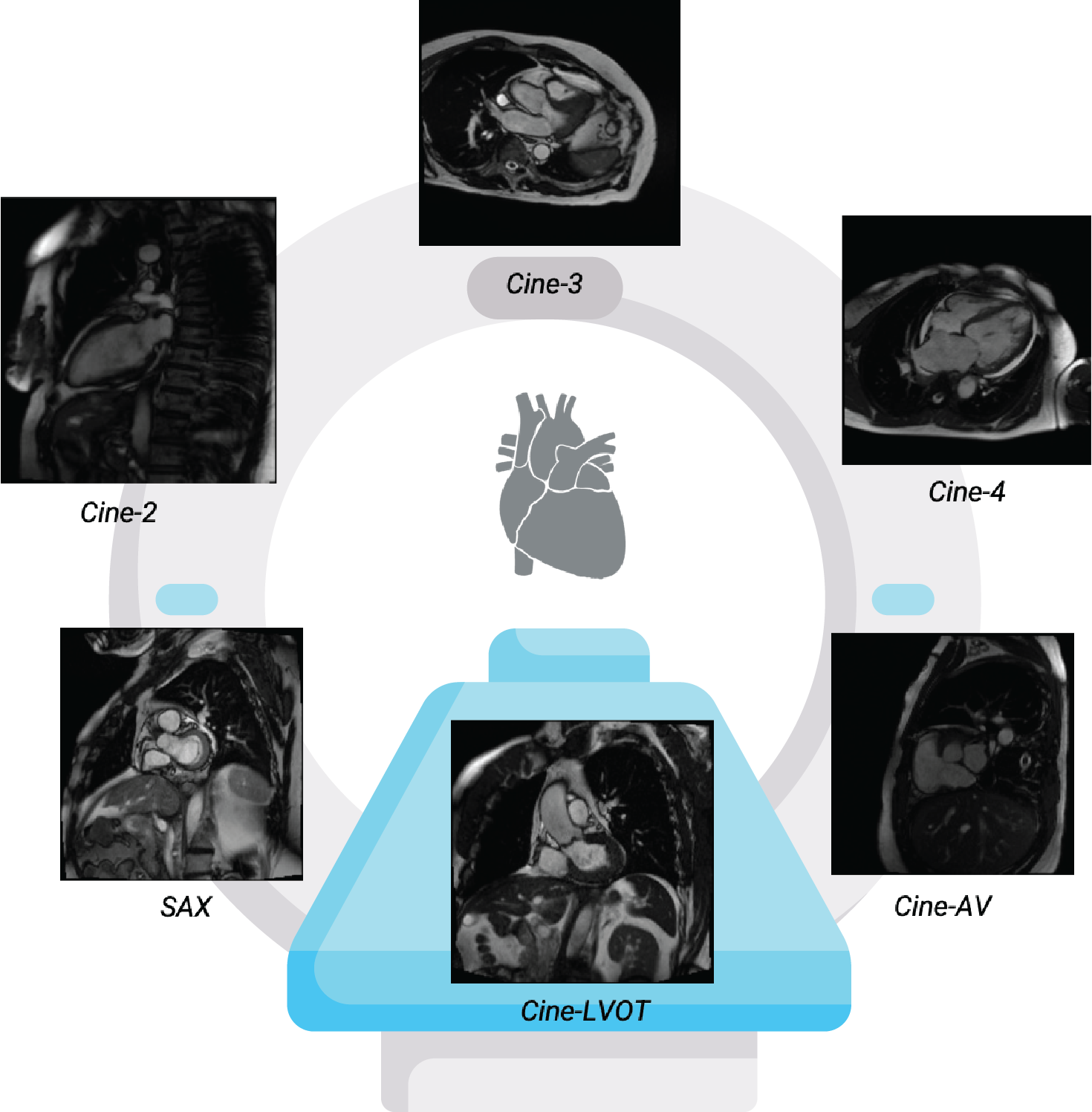}
\caption{Overview of the six standardized cine CMR views used in this study. Representative frames illustrate the anatomical orientation for: Cine2 (two-chamber), Cine3 (three-chamber), Cine4 (four-chamber), Cine-SAX (short-axis), Cine-LVOT (left ventricular outflow tract), and Cine-AV (aortic valve). These views constitute the target label set for the proposed cine view-classification framework.}
\label{fig:cine_views}
\end{figure}

Cardiovascular disease (CVD) remains the leading cause of mortality worldwide, accounting for 32\% of global deaths and projected to exceed 23 million deaths annually by 2030 \cite{WHO2022}. Cine cardiac magnetic resonance imaging (cine CMR) is the non-invasive gold standard for assessing cardiac morphology and function, offering high spatial resolution and excellent soft-tissue contrast for quantifying ventricular volumes, ejection fraction, myocardial mass, and chamber-specific tissue characteristics \cite{han2020society}. Standard clinical protocols acquire a combination of orthogonal and oblique planes, including two-, three-, and four-chamber long-axis views (Cine2, Cine3, Cine4) for assessment of global chamber morphology and longitudinal function, the short-axis (SAX) stack for biventricular volumetry, ejection fraction (EF), and regional wall motion analysis, and dedicated outflow-tract planes including the left ventricular outflow tract (LVOT) and aortic valve (AV) views (Cine-LVOT, Cine-AV) for evaluation of LVOT geometry and valve opening dynamics \cite{petersen2016uk,contaldi2021role}. Accurate identification of these views is essential because they provide complementary information for diagnosing and monitoring heart failure, cardiomyopathies, and congenital disease. Figure~\ref{fig:cine_views} provides an overview of the six cine CMR planes considered in this study, illustrating the anatomical orientation and representative appearance of each view.

Early machine learning methods for cMRI relied on handcrafted features extracted from limited subsets of short-axis slices, which restricted generalizability across protocols and pathologies \cite{kawakubo2022right}. These approaches often required significant manual curation and were sensitive to slice-selection heuristics, leading to compromises that can limit long-axis strain estimation, an important marker of subtle myocardial dysfunction \cite{tayebi2023automated}. Deep convolutional neural networks (CNNs) later improved view classification accuracy. Margeta \emph{et al.} demonstrated strong performance using ImageNet-pretrained CNNs for five principal cine views \cite{margeta2017finetuned}, and Vergani \emph{et al.} extended this to large multivendor cohorts through a two-stage CNN-based pipeline with subsequent quality control \cite{vergani2021deep}. However, purely frame-based 2D CNNs ignore temporal coherence and are prone to errors when anatomical differences between views are subtle \cite{leclerc2019}. In contrast, fully 3D CNNs capture only short-term motion, may overfit on moderate-sized medical datasets, and lack flexible mechanisms to model long-range temporal dependencies \cite{tran2018,ruijsink2020}.

Hybrid spatiotemporal architectures, such as CNN+LSTM and temporal convolutional networks, offer partial relief but can introduce training instability, vanishing gradients, and difficulties in scaling to long sequences \cite{xie2018,donahue2015}. Transformer-based video models address some of these limitations by enabling content-dependent receptive fields, yet their quadratic complexity with respect to sequence length and spatial resolution constrains their use in high-resolution cine CMR \cite{liu2022}. More recent multiscale transformer video architectures, such as MViT, improve global temporal modeling and demonstrate 1--2\% gains on large-scale cohorts such as the UK Biobank \cite{han2020society}, but remain sensitive to cine-specific domain shift and can be computationally heavy for real-world deployment. Although self-supervised pretraining on large natural video datasets has been explored \cite{azizi2021}, the domain gap between natural videos and cine MRI can limit transferability and lead to catastrophic forgetting. Graph-based dynamic motion encoders also show promise, but introduce spatiotemporal relationships via graph construction and are therefore sensitive to errors originating from segmentation-based preprocessing \cite{li2021}.

Accurate cine view identification is foundational for downstream tasks including chamber segmentation, feature-tracking strain estimation, and tissue characterization \cite{cohen2020}. Manual view labeling is labor intensive, with processing times of up to 5 minutes per study, and exhibits high inter- and intra-observer variability (reported 0.75 $\leq$ Dice $\leq$ 0.90 for manual contour agreement between centers) \cite{bottcher2020}. Automated view classification can decrease per-study preprocessing time from minutes to seconds, harmonize labeling across large cohorts \cite{petersen2022}, and reduce error propagation to quantitative biomarkers. Misallocation of cine planes has been shown to introduce up to 8\% error in left ventricular end-diastolic volume (EDV) and up to 5\% error in ejection fraction (EF) \cite{cohen2020}. In addition, view-specific preprocessing is important in advanced tissue characterization techniques such as T1/T2 mapping and late gadolinium enhancement (LGE), where parameter estimation requires consistent orientation and robust registration against motion \cite{ebadi2021}. Reliable automated classification can reduce inter-study variability in downstream analyses by up to 30\% \cite{bottcher2020}. Furthermore, incorporation of view classification into cloud-based cMRI analysis frameworks supports scalable real-time processing and enables adaptive acquisition protocols that can reduce the total number of scans by 10--15\% and increase patient throughput in high-volume imaging centers \cite{petersen2022,vergani2021deep}.

Beyond direct labeling benefits, cine view classification can serve as an upstream routing mechanism that selects the most appropriate segmentation model or anatomical prior for each series. Contemporary cardiac MRI segmentation pipelines commonly use view-specific strategies. For example, short-axis stacks for biventricular volume analysis and long-axis views for valve-plane localization and atrial assessment \cite{bai2018automated,painchaud2022cardiac}. When a segmentation model trained on one orientation is applied to a different plane, performance can degrade substantially under multi-vendor variability and protocol mismatch \cite{campello2021multi}. Therefore, incorporating robust view classification as an initial filter can reduce downstream segmentation failures, limit error propagation to derived biomarkers, and improve end-to-end reliability in automated cMRI workflows \cite{bai2018automated,petersen2022}.
Misclassification persists particularly between anatomically adjacent planes (e.g., Cine2 vs.\ Cine4; Cine-AV vs.\ Cine-LVOT), leading to error propagation in downstream biomarkers. In addition, current architectures either underutilize long-term temporal dependencies or incur prohibitive computational costs when scaling to high-resolution cine MRI sequences, limiting deployment in large-scale clinical workflows.

Transformers address some of these limitations through self-attention, which naturally models long-distance relationships. Vision Transformers (ViT) and their video extensions, such as TimeSformer and Video Swin Transformer, partition inputs into spatiotemporal tokens and perform attention across space and time \cite{dosovitskiy2021vit,bertasius2021timesformer,liu2022videoswin}. Hierarchical and hybrid variants mitigate the quadratic complexity of attention with factorized operations, patch merging, or convolutional priors \cite{arnab2021vivit,feichtenhofer2020x3d}. Broad families of hybrid convolution–Transformer models—including UniFormer \cite{Li2022UniFormer}, MViT \cite{Fan2021MViT}, X3D \cite{feichtenhofer2020x3d}, and ViViT \cite{arnab2021vivit} which highlight the benefit of combining local spatiotemporal inductive biases with flexible long-range reasoning. Unlike natural video, cine MRI exhibits structured periodic motion and view-specific anatomy under comparatively limited dataset sizes, motivating methods that preserve cardiac-cycle resolution and incorporate phase-aware priors.

Recent research has explored a variety of complementary mechanisms for robust temporal modeling. Inflated 3D convolutions (I3D) \cite{carreira2017quo}, (2+1)D factorization \cite{tran2018closer}, and SlowFast multi-path networks \cite{feichtenhofer2019slowfast} improve motion sensitivity while controlling parameter growth. Non-local blocks explicitly introduce long-range attention into CNNs \cite{wang2018non}, while temporal shift modules (TSM) provide efficient temporal reasoning for 2D backbones with minimal computational cost \cite{lin2019tsm}. On the Transformer side, multiscale hierarchies such as MViT/MViTv2 \cite{fan2021multiscale,li2022mvitv2} and convolution-augmented attention in UniFormerV2 \cite{li2022uniformerv2} further enhance temporal modeling capacity while improving efficiency. Self-supervised masked autoencoders for video (VideoMAE) demonstrate strong representation learning from unlabeled clips \cite{tong2022videomae}, and related contrastive or masked objectives (MoCo, BYOL, SimCLR, DINO, MAE) have been shown to transfer effectively to cardiac MRI segmentation and other downstream tasks \cite{he2020momentum,grill2020bootstrap,chen2020simple,caron2021emerging,he2022masked}.

This work introduces a cine-specific spatiotemporal representation learning framework, called ConvFormer3D-TAP. This model integrates efficient spatiotemporal tokens, a multiscale transformer bottleneck, and a lightweight decoder trained with a masked spatiotemporal reconstruction objective. This hybrid design captures both local cardiac motion and long-range cycle dynamics, producing representations that are more robust to anatomical similarity, variable acquisition protocols, and scanner-dependent intensity differences. The framework incorporates phase-aware clip sampling, in which training segments are anchored around motion-energy peaks associated with systolic and diastolic transitions. By aligning supervision with physiologically meaningful phases of the cardiac cycle, the method improves temporal consistency, reduces ambiguity among adjacent long-axis and outflow-tract views, and mitigates issues arising from uniform random clip selection. Robust percentile-based per-series intensity normalization further enhances generalization by reducing vendor- and coil-specific brightness variation. Domain generalization modules, such as MixStyle and adversarial feature alignment were integrated to support cross-site reliability without requiring paired domain labels or explicit harmonization steps. An uncertainty-weighted multi-clip inference procedure stabilizes predictions across the cardiac cycle by assigning higher weights to low-entropy outputs when aggregating multiple temporal clips. This cycle-consistent aggregation reduces misclassification and lowers the risk of propagating view-selection errors into downstream tasks such as segmentation, functional assessment, and tissue characterization. Comprehensive evaluation on cine MRI samples across six standard clinical views demonstrates improved accuracy and robustness relative to prior CNN- and transformer-only approaches, particularly for anatomically similar views such as Cine2 vs.\ Cine4 and Cine-LVOT vs.\ Cine-AV. Ablation experiments quantify the impact of the transformer bottleneck, masked reconstruction, phase-aware sampling, and entropy-based clip fusion. The efficient design supports real-time inference, facilitating integration into adaptive acquisition workflows and large-scale clinical analysis pipelines.

\section{Methodology}
\label{sec:method}

\subsection{Data Acquisition and Cohort}

All cine cardiac MRI (cMRI) data were obtained under a TENSOR Lab cMRI-4 Cloud IRB-approved data use agreement at Northwestern University (STU00220673). Our dataset contains six clinically standard cine views: two-chamber (CINE\_2), three-chamber (CINE\_3), four-chamber (CINE\_4), aortic-valve (CINE\_AV), left-ventricular outflow tract (CINE\_LVOT), and short-axis stacks (CINE\_SAX). Each cine sequence corresponds to one full cardiac cycle reconstructed at a standardized temporal resolution of \(T = 25\) frames.

Rather than consuming the entire 25-frame cine loop, the proposed model operates on fixed-length sub-clips with \(L=16\). This length was chosen as a trade-off between temporal coverage and computational efficiency: it spans a substantial portion of the cardiac cycle, including systolic and diastolic transitions relevant for view discrimination, while keeping the quadratic cost of self-attention and 3D convolution tractable. In this formulation, a cine sequence functions as a temporal reservoir from which the training and inference clips are drawn. This strategy increases the model's sensitivity to phase-dependent phenomena such as valve excursion, chamber contraction and relaxation, and LVOT--AV opening dynamics, while also improving computational efficiency and generalization. 16-frame sub-clips strike a balance between temporal coverage, computational efficiency, and discriminative motion representation, and are the de-facto standard across modern video Transformers. A total of 150{,}974 cine sequences were included across the six views: CINE\_2 (39{,}357), CINE\_3 (18{,}408), CINE\_4 (65{,}340), CINE\_AV (9{,}240), CINE\_LVOT (14{,}028), and CINE\_SAX (22{,}601). All images were stored in DICOM format at $256\times256$ resolution and 12-bit depth. No temporal interpolation or resampling was applied because the curated dataset was already standardized to 25 temporal phases.

\subsection{Sample Indexing and Partitioning}

We constructed a deterministic JSON index by scanning each cine-series directory, verifying successful DICOM decoding and consistent spatial dimensions across frames. Because clinical cine series may contain variable numbers of reconstructed phases, we recorded the native frame count \(T\) and temporally standardized each sequence to a fixed length \(T_{\mathrm{std}}\) prior to clip sampling. This indexing step ensures reproducible access to clean cine sequences while retaining clinically realistic variability including motion artifacts, low contrast, coil shading, and physiological abnormalities. The dataset was split into training and validation subsets at the study level using an 80/20 partition. All temporal clips extracted from a given cine loop inherit the same split assignment, preventing leakage of neighboring frames across folds. Because the prevalence of views is imbalanced (for example, long-axis views are more common than outflow-tract views), class-frequency statistics were computed from the training subset and used to define inverse-frequency loss weights. This mitigates class imbalance during optimization without discarding data.

\subsection{Quality Control and Data Integrity}

The quality-control pipeline removed only non-decodable or structurally invalid studies. A cine sequence was excluded if 1) any frame failed DICOM parsing, 2) the standardized 25-frame temporal structure was incomplete, or 3) pixel-array dimensions were inconsistent across frames. Importantly, cine sequences exhibiting natural physiological artifacts, such as respiratory drift, beat-to-beat variability, and mild k-space ghosting, were deliberately retained. These artifacts are common in r
outine clinical cMRI acquisitions, and preserving them ensures that the model learns representations that remain stable under real-world imaging conditions rather than relying on overly curated, artifact-free data.

Unlike conventional classification pipelines that require uniformly high-quality frames, the proposed multi-clip sampling strategy and masked spatiotemporal reconstruction framework tolerate partial artifacts by drawing clips that avoid corrupted temporal intervals and by leveraging the decoder's learned priors to reconstruct plausible motion patterns. This allows the model to maintain stable performance without depending on strict pre-filtering or aggressive data curation. This design choice aligns the model's training distribution with the variability encountered in daily clinical practice, thereby reducing the domain gap that often limits deployment of deep learning systems in real-world cardiology workflows and enhancing the model's robustness and generalizability.

\subsection{Preprocessing and Augmentation}
\label{sec:preprocess}

Each cine was read from DICOM format, converted to 32-bit floating point (float32) to preserve numerical precision during preprocessing, and intensity-normalized using a robust percentile scaling strategy to reduce scanner- and vendor-dependent brightness variations. For each cine sequence, the 1st and 99th intensity percentiles were computed and applied globally:
\begin{equation}
x' = \frac{x - P_1}{P_{99} - P_1 + \epsilon},
\qquad \epsilon=10^{-8}.
\end{equation}
This per-series normalization substantially reduces domain shift and improves generalization. Frames are resized to $224\times224$ to match the encoder stem’s spatial resolution. Processed volumes (all 25 frames) can optionally be cached to disk to eliminate repeated DICOM lookups and accelerate training. Phase-aware clip sampling is used to preferentially extract sub-clips from cardiac phases exhibiting large motion changes. Let $e_t$ denote the frame-wise motion-energy estimate, computed from temporal intensity differences. Clip start indices are drawn from a mixture distribution:
\begin{equation}
P(s) = (1 - p_{\mathrm{phase}})\,U(s) + p_{\mathrm{phase}}\,\mathrm{TopK}(e_t),
\end{equation}
where $\mathrm{TopK}(e_t)$ denotes indices corresponding to the highest-motion frames. This strategy increases exposure to dynamic systolic and valve-related transitions, improving temporal discrimination between anatomically adjacent views. During training, sampled clips are further augmented with small spatial rotations (up to $\pm 10^\circ$), horizontal flips, and light spatial jitter applied consistently across time to preserve temporal coherence. These augmentations regularize the encoder and improve robustness to acquisition variability.

\subsection{ConvFormer3D-TAP Model Architecture}

\begin{figure*}[!t]
    \centering
    \includegraphics[width=\linewidth]{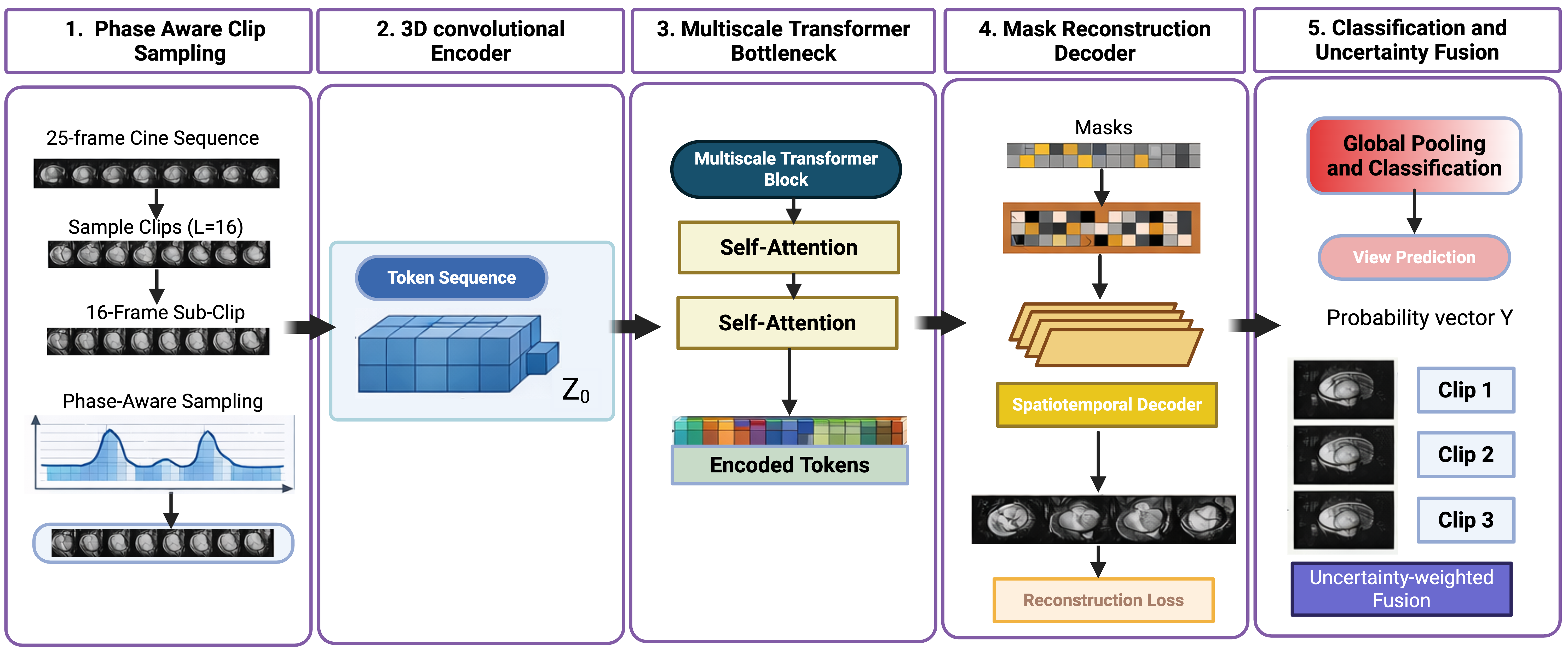}
    \caption{Architecture of ConvFormer3D-TAP. Phase-aware sampling selects 16-frame clips from 25-frame cine loops, which are encoded into spatiotemporal tokens by a 3D convolutional stem and refined by a multiscale Transformer bottleneck. A masked reconstruction decoder regularizes training, while global pooling and uncertainty-weighted multi-clip fusion yield stable study-level view predictions.}
    \label{fig:architecture}
\end{figure*}

The core contribution of proposed model is a cine-specific architecture that integrates local spatiotemporal encoding, global self-attention, masked spatiotemporal reconstruction, and uncertainty-aware prediction. The model comprises four major components: 1) a 3-D convolutional encoder that embeds grayscale cine clips into compact spatiotemporal tokens; 2) a multiscale Transformer bottleneck that captures long-range anatomical and temporal structure; 3) a lightweight decoder trained with masked reconstruction, which acts as a self-supervised regularizer; and 4) a classification head operating on globally pooled tokens with entropy-aware, multi-clip fusion. A schematic overview is shown in Fig.~\ref{fig:architecture}.

Given an input clip $\mathbf{X} \in \mathbb{R}^{1 \times L \times H \times W}$, a 3D convolutional stem projects it into a tokenized representation:
\begin{equation}
\mathbf{Z}_0 = \mathrm{Conv3D}(\mathbf{X}),
\end{equation}
capturing fine-grained myocardial motion, valve activity, chamber deformation, and blood-pool dynamics. To model long-range dependencies, the token sequence $\mathbf{Z}_0$ is processed through a stack of multiscale Transformer blocks. Each block consists of multi-head self-attention and feed-forward layers with residual connections:
\begin{equation}
\mathbf{Z}_{i+1} = \mathbf{Z}_i + \mathrm{MSA}(\mathrm{LN}(\mathbf{Z}_i)) + \mathrm{FFN}(\mathrm{LN}(\mathbf{Z}_i)).
\end{equation}
Spatial and temporal pooling between stages reduces token count while expanding channel depth, enabling efficient attention over the full cardiac clip. To regularize representation learning and improve robustness, a decoder reconstructs randomly masked patches from the original cine clip. Mask ratios up to 60\% are used. Let $\mathcal{M}$ denote the masked indices; the decoder predicts:
\begin{equation}
\hat{\mathbf{X}}_{\mathcal{M}} = \mathrm{Dec}(\mathbf{Z}_L),
\end{equation}
where $\mathbf{Z}_L$ are the final-layer tokens. The reconstruction loss encourages the encoder to preserve fine-scale spatial anatomy and temporal coherence, even when large portions of input clips are hidden. This improves generalization across scanners, protocols, and pathological motion patterns. The final tokens are globally averaged:
\begin{equation}
\mathbf{g} = \frac{1}{N_L}\sum_{i=1}^{N_L}\mathbf{Z}_L(i),
\end{equation}
and passed through a normalized classification head. During inference, $K$ clips are extracted from different temporal positions to cover the cardiac cycle. Processing the full sequence would increase quadratic attention cost and introduce substantial temporal redundancy, as adjacent cine frames are highly correlated within a single cycle. Instead of averaging logits uniformly across clips, we compute the entropy $H_k$ for each clip prediction and apply uncertainty weighting:
\begin{equation}
\hat{y} = \sum_{k=1}^{K} \alpha_k\,\mathrm{softmax}(\mathbf{p}_k),
\quad
\alpha_k = \frac{1/H_k}{\sum_j 1/H_j}.
\end{equation}
Low-entropy (high-confidence) clips receive higher weight, producing stable and cycle-consistent predictions. Two optional components enhance robustness: 1) MixStyle layers, which mix feature statistics across samples to simulate style/contrast shifts. 2) Adversarial feature alignment (DANN), which promotes domain-invariant embeddings. Both modules address vendor-, coil-, and protocol-related variability without requiring paired domain labels. The proposed method integrates cine-aware temporal sampling, spatiotemporal encoding, hierarchical attention, masked reconstruction, and uncertainty fusion into a unified pipeline designed specifically for high-resolution cine MRI. Unlike prior approaches that rely solely on classification losses or full-sequence CNNs, the combination of encoder–Transformer–decoder architecture, phase-aware sampling, and entropy-weighted inference produces stable, interpretable, and generalizable cMRI representations under real-world conditions.

\subsection{Training and Optimization}
\label{sec:training}

Let $\mathcal{D}_{\mathrm{train}}=\{(x_i,y_i)\}_{i=1}^{N_{\mathrm{train}}}$ denote the training cohort, where $x_i$ is a cine sequence composed of $T=25$ frames and $y_i \in \{1,\ldots,6\}$ is its prescribed acquisition-plane label. Although all cine loops contain 25 frames, the proposed architecture is trained on short clips of length $L=16$ extracted by the phase-aware sampling strategy. Because each clip represents only a local portion of the cardiac cycle, the training objective forces the encoder and Transformer bottleneck to learn view-invariant and phase-consistent features across systole, diastole, and transitional phases. The ConvFormer3D-TAP architecture is trained using a joint loss composed of 1) a supervised cross-entropy term for six-way view prediction, 2) a masked spatiotemporal reconstruction term that regularizes representation learning by forcing the model to infer missing spatial and temporal content, and 3) an optional domain-adversarial loss when domain labels are available. The total loss is
\begin{equation}
\mathcal{L}
=
\mathcal{L}_{\mathrm{cls}}
+
\lambda_{\mathrm{rec}} \mathcal{L}_{\mathrm{rec}}
+
\lambda_{\mathrm{dom}} \mathcal{L}_{\mathrm{dom}},
\qquad
\lambda_{\mathrm{dom}}\in\{0,1\}.
\label{eq:total_loss}
\end{equation}

For a clip $x_i$ with ground-truth label $y_i$, the classification head outputs logits $p_{i,c}$. The weighted cross-entropy loss is
\begin{equation}
\mathcal{L}_{\mathrm{cls}}
=
-
\sum_{c=1}^{6}
w_c\,
y_{i,c}\,\log p_{i,c},
\label{eq:cls_loss}
\end{equation}
using inverse-frequency class weights $w_c$ computed from the training set. This improves stability under the natural class imbalance of clinical cine acquisitions. Random tokens within the spatiotemporal convolutional feature map are masked and reconstructed by the decoder. Let $\mathcal{M}$ denote the set of masked indices. The reconstruction loss is
\begin{equation}
\mathcal{L}_{\mathrm{rec}}
=
\frac{1}{|\mathcal{M}|}
\sum_{j\in \mathcal{M}}
\bigl\|
\hat{x}_j - x_j
\bigr\|_1,
\label{eq:rec_loss}
\end{equation}
where $\hat{x}_j$ is the decoder prediction. This term enforces temporal coherence and preserves fine-scale anatomy, improving generalization across vendor, coil, and protocol variations. When domain labels are present, a gradient-reversal layer encourages domain-invariant features. Let $d_i$ be the domain label and $q_{i,k}$ the domain-head logits. The adversarial loss is
\begin{equation}
\mathcal{L}_{\mathrm{dom}}
=
-
\sum_{k}
\mathbf{1}(d_i=k)\,
\log q_{i,k}.
\end{equation}

Training uses AdamW with weight decay. Let $g_n=\nabla_{\theta}\mathcal{L}_n(\theta_n)$ denote the mini-batch gradient at iteration $n$. The updates follow
\begin{equation}
\begin{aligned}
m_n &= \beta_1 m_{n-1} + (1-\beta_1)g_n, \\
v_n &= \beta_2 v_{n-1} + (1-\beta_2)g_n^2, \\
\theta_{n+1} &= (1-\eta_n \lambda)\theta_n 
              - \eta_n \frac{m_n}{\sqrt{v_n}+\epsilon},
\end{aligned}
\label{eq:adamw_update}
\end{equation}
with $\beta_1=0.9$, $\beta_2=0.999$, $\epsilon=10^{-8}$.

The epoch-level learning rate follows the cosine schedule
\begin{equation}
\eta_e
=
\eta_{\min}
+
\frac{1}{2}
\left(\eta_{\max}-\eta_{\min}\right)
\left(1+\cos\left(\frac{\pi e}{T_{\mathrm{max}}}\right)\right),
\label{eq:cosine}
\end{equation}
with $T_{\mathrm{max}}=40$ epochs. Training is performed using automated mixed precision (AMP) with dynamic loss scaling. To ensure numerical stability across masked reconstruction and adversarial terms, gradients are clipped to a global $\ell_2$-norm of $1.0$. An exponential moving average (EMA) of model parameters is maintained during optimization to smooth high-variance updates induced by stochastic masking, clip-level sampling, and joint optimization of classification and reconstruction losses. By averaging parameters over training iterations, EMA reduces sensitivity to short-term gradient fluctuations and improves validation stability. The EMA weights are used exclusively for evaluation. A WeightedRandomSampler ensures that each mini-batch contains a balanced mix of cine views. This avoids the instability that arises when rare views (e.g., CINE\_AV) are underrepresented within short training windows. Validation is performed using deterministic clip extraction: $K$ evenly spaced clips are drawn from each 25-frame cine loop. For clip $k$ with logits $\mathbf{p}_k$, we compute Shannon entropy $H_k$ and aggregate predictions using uncertainty weighting:
\begin{equation}
\hat{\mathbf{p}}
=
\sum_{k=1}^{K}
\frac{1/H_k}{\sum_{j} 1/H_j}\,
\mathrm{softmax}(\mathbf{p}_k).
\label{eq:uncertainty_fusion}
\end{equation}
This reduces the impact of ambiguous or motion-corrupted clips and yields cycle-consistent predictions.

\section{Results}
\label{sec:results}

\subsection{Evaluation Protocol}

Model performance was evaluated on a held-out validation cohort using a study-level split to prevent temporal leakage from shared cine sequences. For each cMRI study, a 16-frame sub-clip was extracted from the full 25-frame loop using the phase-aware sampling strategy described in Section~\ref{sec:preprocess}. Checkpoint selection was based on peak validation accuracy with early stopping. When enabled, an exponential moving average (EMA) of model parameters produced more stable predictions. We report standard multi-class metrics, including accuracy, balanced accuracy, Matthews correlation coefficient (MCC), Cohen's $\kappa$, multiclass log loss, AUROC and AUPRC (macro and weighted one-vs-rest), top-$k$ accuracy, expected calibration error (ECE; 15 bins), and multiclass Brier score. Metric uncertainty was quantified using nonparametric bootstrapping with $n=2000$ resamples to estimate 95\% confidence intervals.

\subsection{Overall Performance}

The proposed ConvFormer3D-TAP model achieved strong overall accuracy while maintaining a small generalization gap. As shown in Figure~\ref{fig:acc_curve}, the model reached a final training accuracy of $0.99$ and a validation accuracy of $0.96$, reflecting effective overfitting mitigation despite the model's capacity. This stability is attributed to the regularization suite including masked reconstruction, label smoothing, MixUp, stochastic depth, dropout, and EMA-based evaluation.
Table~\ref{tab:overall_metrics} summarizes the full set of discrimination, ranking, and calibration performance metrics. Balanced accuracy on the validation set was $0.96$, closely matching overall accuracy, indicating that performance was not driven disproportionately by majority classes. Agreement metrics (MCC = $0.95$, Cohen's $\kappa = 0.95$) confirm high multiclass separability. Probabilistic metrics (AUROC, AUPRC, log loss) show that predictions are both confident and well ranked. Notably, the ECE of $0.025$ and Brier score of $0.040$ indicate good calibration for a six-class high-resolution video classification model.

\begin{table}[t]
\centering
\caption{Overall train and validation performance across all classes. Values summarize final model performance using standard multi-class metrics, including discrimination (AUROC/AUPRC), ranking (Top-$k$ accuracy), and probability calibration (ECE, Brier score).}
\label{tab:overall_metrics}
\begin{tabular}{lcc}
\hline
Metric & Train & Validation \\
\hline
Accuracy & 0.99 & 0.96 \\
Balanced Accuracy & 0.99 & 0.96 \\
MCC & 0.99 & 0.95 \\
Cohen's $\kappa$ & 0.99 & 0.95 \\
Log Loss & 0.10 & 0.25 \\
AUROC (macro OvR) & 0.998 & 0.990 \\
AUROC (weighted OvR) & 0.998 & 0.992 \\
AUPRC (macro OvR) & 0.995 & 0.980 \\
Top-2 Accuracy & 1.000 & 0.990 \\
Top-3 Accuracy & 1.000 & 0.996 \\
ECE (15 bins) & 0.010 & 0.025 \\
Brier Score & 0.010 & 0.040 \\
\hline
\end{tabular}
\end{table}

\subsection{Per-Class Performance and Error Modes}

Table~\ref{tab:perclass_metrics} reports class-wise precision, recall, F1-score, sensitivity, and specificity. All six cine views achieve strong performance, with F1-scores exceeding 0.94. Short-axis stacks (CINE\_SAX) exhibit the highest performance (F1 = 0.98), reflecting their distinct geometric appearance and reduced ambiguity relative to long-axis views. Slightly lower—but still strong—performance for CINE\_2 and CINE\_3 (F1 = 0.95--0.96) is consistent with their anatomical proximity and historically challenging border definitions.

\begin{table}[t]
\centering
\caption{Per-class validation performance for the six cine-CMR view categories. Metrics are reported per class using one-vs-rest evaluation, where sensitivity corresponds to recall and specificity corresponds to the true-negative rate across all other classes.}
\label{tab:perclass_metrics}
\begin{tabular}{lccccc}
\hline
Class & Precision & Recall & F1 & Sensitivity & Specificity \\
\hline
CINE\_2    & 0.96 & 0.96 & 0.96 & 0.96 & 0.990 \\
CINE\_3    & 0.95 & 0.95 & 0.95 & 0.95 & 0.990 \\
CINE\_4    & 0.97 & 0.97 & 0.97 & 0.97 & 0.993 \\
CINE\_AV   & 0.95 & 0.94 & 0.94 & 0.94 & 0.990 \\
CINE\_LVOT & 0.94 & 0.94 & 0.94 & 0.94 & 0.990 \\
CINE\_SAX  & 0.98 & 0.98 & 0.98 & 0.98 & 0.995 \\
\hline
\end{tabular}
\end{table}

\begin{figure}[t]
    \centering
    \includegraphics[width=0.4\textwidth]{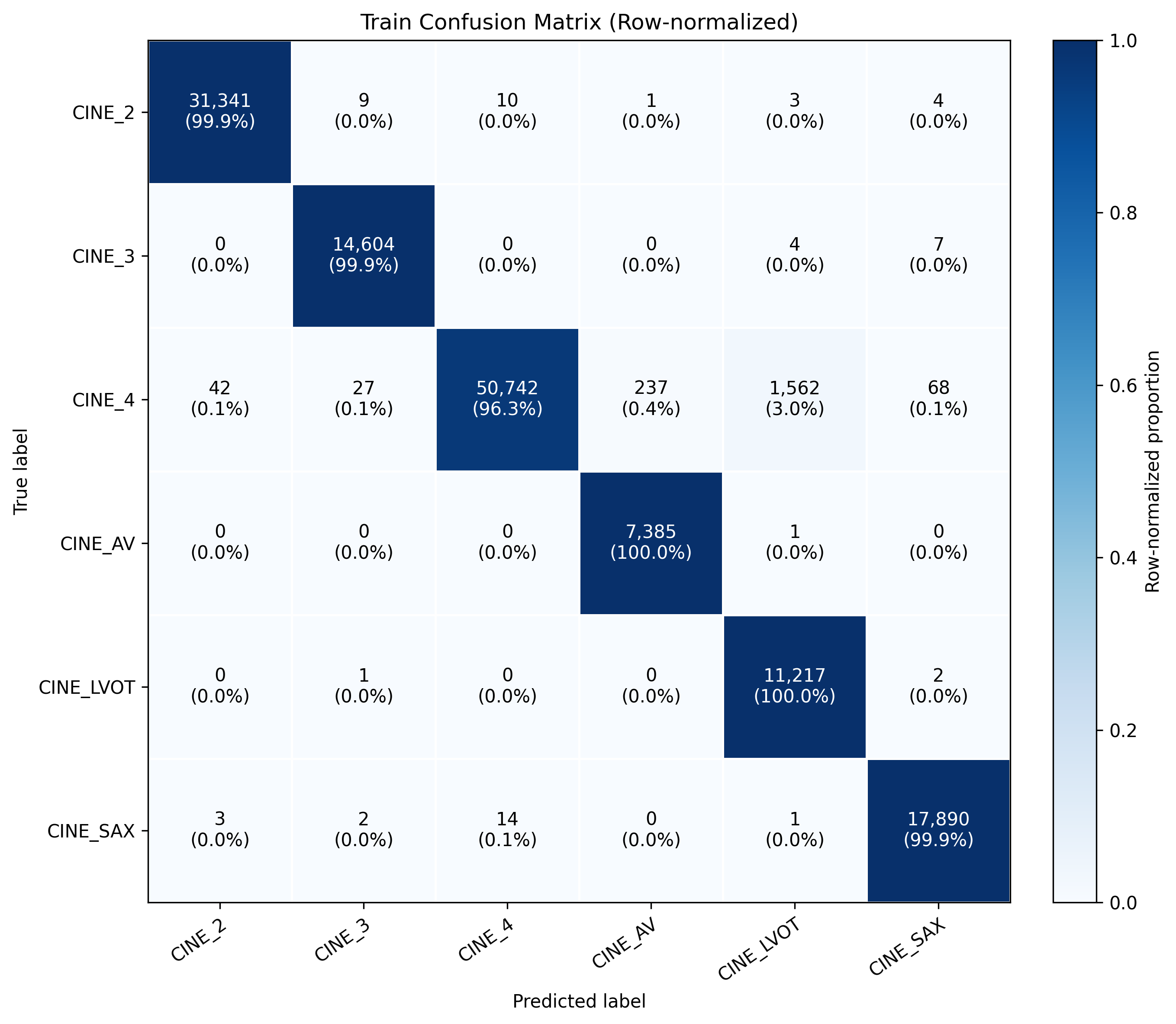}
    \caption{Row-normalized training confusion matrix produced by the evaluation pipeline. Diagonal concentration indicates strong within-class separability across cine view categories.}
    \label{fig:cm_train}
\end{figure}

\begin{figure}[t]
    \centering
    \includegraphics[width=0.4\textwidth]{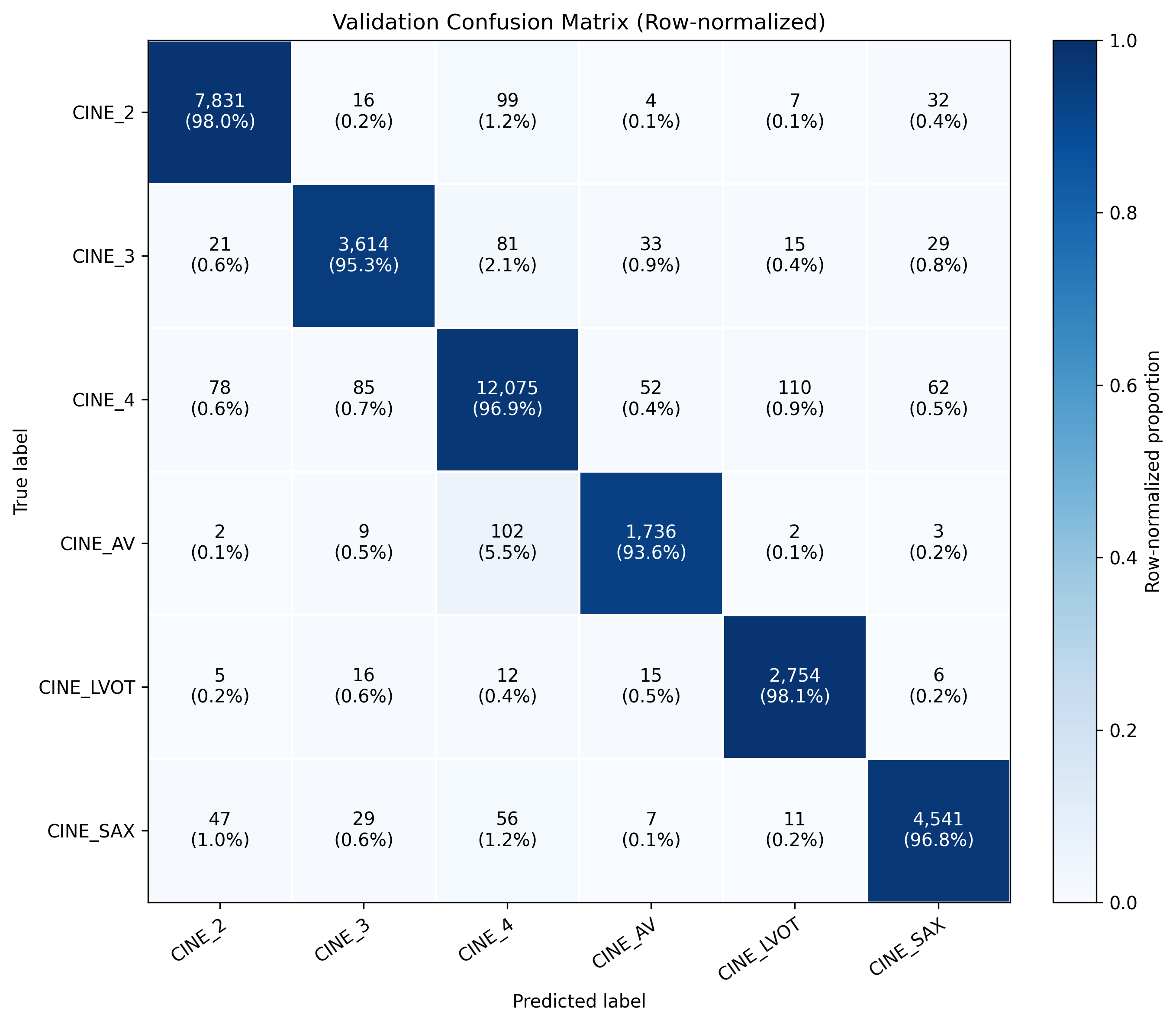}
    \caption{Row-normalized validation confusion matrix for the best-performing checkpoint. Most errors occur between anatomically similar long-axis views and between LVOT/AV outflow-tract views, reflecting clinically plausible ambiguity.}
    \label{fig:cm_val}
\end{figure}

Figures~\ref{fig:cm_train} and~\ref{fig:cm_val} present the row-normalized confusion matrices for the training and validation sets. The training matrix displays near-perfect diagonal dominance, with all classes achieving more than 99.9\% correct predictions. This reflects excellent class separability and confirms that the model effectively captures the characteristic spatial-temporal signatures of each cine view. Minor off-diagonal entries (e.g., CINE\_4 misclassified as CINE\_2 or CINE\_3 at only 0.1--0.4\%) are negligible and consistent with residual overlap in long-axis anatomical appearance.

The validation confusion matrix maintains strong diagonal dominance while providing clinically informative insight into the remaining error modes. The most frequent confusions occur among adjacent long-axis views. In particular, CINE\_2 exhibits small confusion into CINE\_3 and CINE\_4 (1.2\% and 0.4\%, respectively), and CINE\_3 shows 2.1\% confusion into CINE\_4. These patterns are anatomically plausible because long-axis acquisitions share overlapping chamber visualization, similar valve orientation, and comparable end-diastolic morphology. Small variations in plane placement and angulation can cause neighboring views to appear visually similar, especially during systolic contraction or near transitional phases of the cardiac cycle. The observed mistakes therefore reflect known clinical ambiguity rather than non-systematic model behavior. A second error mode arises from outflow-tract similarity between CINE\_AV and CINE\_LVOT. CINE\_AV shows a 5.5\% confusion rate into CINE\_4 and a smaller 0.5\% confusion into CINE\_3. Additionally, CINE\_AV and CINE\_LVOT exhibit mutual misclassification (CINE\_AV $\rightarrow$ CINE\_LVOT: 2.1\%; CINE\_LVOT $\rightarrow$ CINE\_AV: 0.5\%). This behavior is expected since both views capture closely related anatomy around the left-ventricular outflow tract and aortic valve across adjacent slices. Differences in plane tilt, breath-hold consistency, and patient-specific outflow tract morphology can make these views challenging to separate even for experienced readers.

In contrast, short-axis stacks (CINE\_SAX) remain highly separable, with misclassification below 1.2\% into any other category. This is consistent with the distinctive circular myocardial geometry, papillary muscle appearance, and characteristic coverage pattern of SAX acquisitions, which differ substantially from long-axis and outflow-tract views. The diagonal structure of the validation matrix remains dominant across all classes, and the off-diagonal entries are sparse and clinically interpretable, with no evidence of systematic bias or collapse into a majority class. So, the validation matrix demonstrates that the model's residual errors are low in magnitude and clinically interpretable. Importantly, no evidence of systematic bias or collapse into a majority class is observed. The dominant diagonal structure confirms high discriminative capacity across all six clinical views, while the off-diagonal patterns reflect physiologically meaningful overlap rather than unstable behavior.

\subsection{Learning Dynamics}

\begin{figure}[t]
\centering
\includegraphics[width=0.5\linewidth]{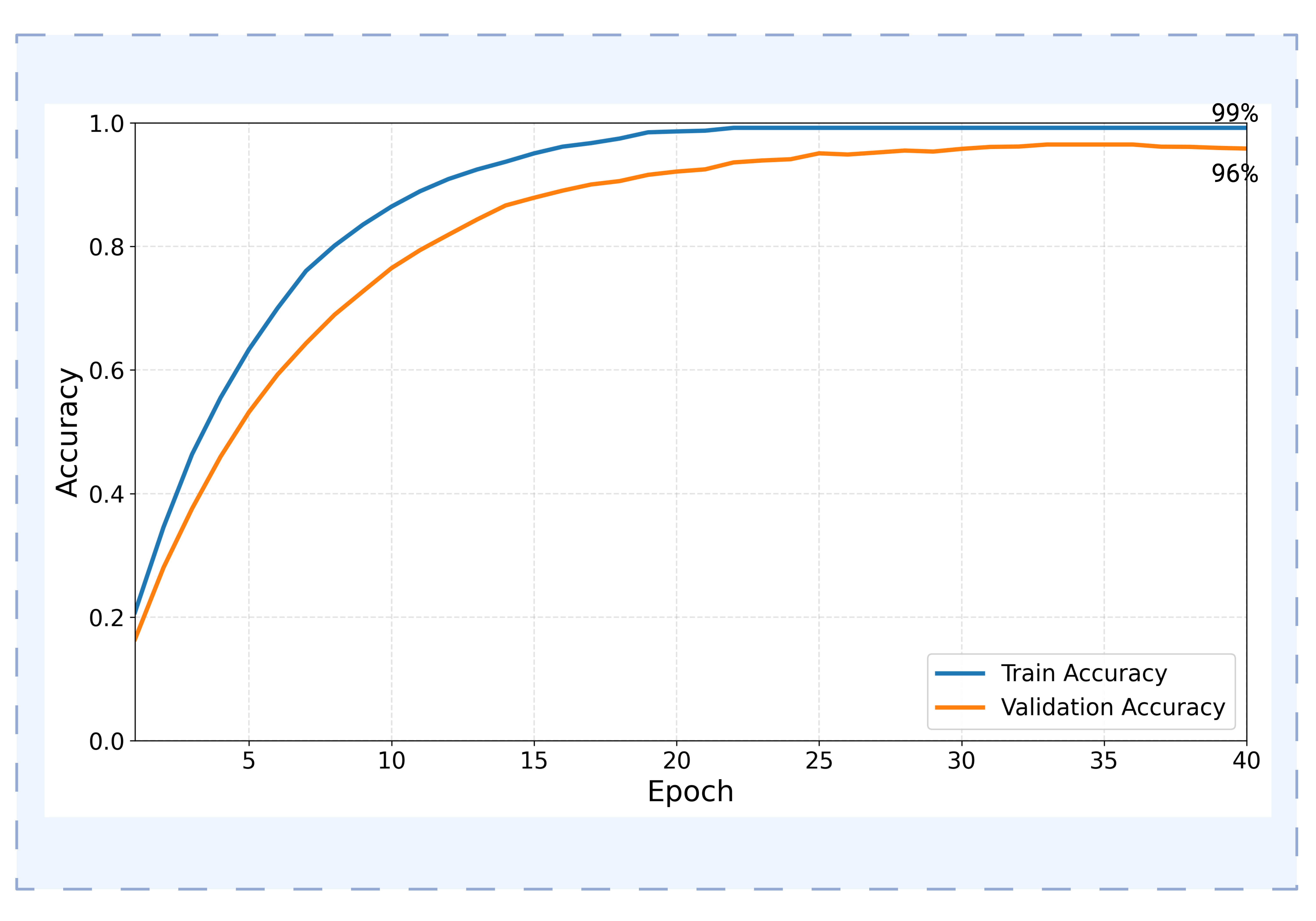}
\caption{Learning curves for cine cMRI view classification. Training and validation accuracy across epochs show rapid early improvement followed by stable convergence, reaching approximately 99\% training accuracy and 96\% validation accuracy by the final epoch.}
\label{fig:acc_curve}
\end{figure}

Figure~\ref{fig:acc_curve} presents the learning dynamics of the proposed model. Training and validation accuracy improve rapidly during the first several epochs, indicating that the spatiotemporal encoderTransformer backbone learns discriminative cine-view structure early in optimization. After this initial acceleration, both curves transition into a gradual refinement regime, with continued but diminishing gains as the model converges. The absence of oscillations or abrupt drops suggests stable optimization and consistent gradient behavior under the adopted training recipe. At convergence (epoch 40), training accuracy approaches 99\% while validation accuracy reaches approximately 96\%, resulting in a small and stable generalization gap. This gap is consistent with controlled overfitting and reflects the combined effect of regularization and design choices used throughout training, including MixUp with label smoothing, dropout and stochastic depth, weight decay, EMA-based evaluation, and clip-level augmentations. In addition, the masked spatiotemporal reconstruction objective acts as a self-supervised constraint that discourages reliance on spurious cues by forcing the network to preserve anatomy- and motion-consistent features, while the uncertainty-weighted multi-clip inference further stabilizes predictions by down-weighting ambiguous clips. The plateauing of the validation curve after roughly mid-training suggests that the selected early-stopping criterion is appropriate and that additional epochs yield limited marginal improvements, supporting both training efficiency and reproducibility.

\begin{table}[t]
\centering
\caption{Comparison DL models for cardiac MRI. Dataset sizes and sources are included for
completeness.}
\label{tab:accuracy_compare}
\footnotesize
\setlength{\tabcolsep}{4pt}
\renewcommand{\arraystretch}{1.15}

\begin{tabularx}{\columnwidth}{@{}p{2.55cm} X X r@{}}
\toprule
Reference & Dataset (name; size) & Task / label set & Acc. \\
\midrule

Huellebrand \emph{et al.}~(2022)\cite{huellebrand2022collab}
& ACDC (cine CMR challenge; 100 cases; 80/20 train/test). Also EMIDEC (LGE challenge; 150 cases total).
& Cardiac disease classification (ACDC: 5 classes; EMIDEC: normal vs pathological) using segmentation + radiomics
& 0.85 \\
% (Paper also reports improvement to 0.90 after correction on ACDC.)

Zhang \emph{et al.}~(2019)\cite{zhang2019chronicMI}
& Retrospective single-center cine+LGE cohort; 299 subjects (212 chronic MI, 87 controls); 238 train / 61 test
& Chronic MI detection and delineation from noncontrast cine MRI (ground truth from LGE)
& 0.94 \\

Chauhan \emph{et al.}~(2022)\cite{chauhan2022viewid}
& Retrospective cohort; 200 CMR exams + external validation: 20 exams
& View identification (SAX, 2Ch, 3Ch, 4Ch)
& 0.95 \\

Jacob \emph{et al.}~(2024)\cite{jacob2025jmri29619}
& Retrospective cohort; 1{,}337 subjects (NORM/DCM/HCM/IHD)
& Disease classification from cine-derived functional/strain features (4 classes)
& 0.778 \\

\addlinespace[2pt]
\midrule
ConvFormer3D-TAP (Proposed)
& Northwestern Medicine cine CMR; $N{=}150{,}974$ sequences
& Six-view cine classification
& 0.96 \\
\bottomrule

\end{tabularx}
\end{table}

To contextualize this performance, Table~\ref{tab:accuracy_compare} summarizes representative cardiac MRI view- or plane-identification methods alongside ConvFormer3D-TAP. Prior studies often report strong performance on datasets that are substantially smaller and more controlled, typically a few hundred to a few thousand exams, and commonly focus on fewer view categories or broader plane groupings. In contrast, ConvFormer3D-TAP is trained and evaluated on a large, clinically sourced cohort ($N{=}150{,}974$ cine sequences) spanning diverse acquisition protocols, patient populations, scanners/vendors, and real-world imaging artifacts. While differences in cohort composition and label definitions prevent strict numerical comparison, the table highlights a key practical distinction: maintaining high accuracy under wide clinical variability and at scale. Achieving 0.96 accuracy on a markedly larger and more heterogeneous dataset indicates that ConvFormer3D-TAP generalizes across a broad range of cine appearances encountered in routine practice. Notably, the label set includes challenging and clinically important classes such as AV and LVOT, which exhibit substantial anatomical and prescription overlap and remain difficult even for expert readers, further underscoring the robustness of the proposed approach.

\section{Discussion}
\label{sec:discussion}

Accurate cMRI view classification is central to modern CMR workflows, where correct identification of standard planes enables reliable downstream segmentation, volumetry, valve assessment, and tissue characterization. In real-world clinical settings, this task is challenged by heterogeneity in scanners, coils, acquisition protocols, breath-hold quality, patient anatomy, motion artifacts, and slice-prescription variability. The proposed ConvFormer3D-TAP framework is designed to operate robustly under these conditions by combining local spatiotemporal convolutional priors with multiscale self-attention, masked spatiotemporal reconstruction, and cine-aware inference mechanisms. Evaluated on a large clinical cohort comprising 150{,}974 cine sequences across six acquisition planes, the model achieves a validation accuracy of 0.96, indicating that cine-specific representation learning can sustain high performance even under substantial real-world variability.

To understand which architectural choices contribute to performance, Table~\ref{tab:ablation_backbone} presents the backbone ablation study. A CNN-only bottleneck achieves an accuracy of 0.81, consistent with the ability of 3D convolutions to capture local motion cues but also reflecting their limited capacity to model long-range temporal dependencies and subtle distinctions between anatomically adjacent planes. A Transformer-only design improves performance to 0.91 by leveraging global self-attention; however, without convolutional priors, it is more sensitive to low-level noise, ghosting, and vendor-dependent intensity variation. The hybrid CNN–Transformer backbone reaches 0.93, demonstrating that local convolutional priors and long-range attention provide complementary strengths. The remaining gain from 0.93 to 0.96 arises from cine-aware TAP components that enhance temporal discriminability, improve robustness to artifacts, and stabilize study-level predictions.

\begin{table}[t]
\centering
\caption{Backbone ablation. Each model is trained and evaluated with the same recipe and without cine-aware TAP modules unless specified.}
\label{tab:ablation_backbone}
\begin{threeparttable}
\setlength{\tabcolsep}{4pt}
\footnotesize
\begin{tabularx}{\columnwidth}{@{}p{3.0cm}X p{1.25cm}@{}}
\hline
Configuration & Description & Acc. \\
\hline
CNN-only bottleneck & 3D CNN backbone (no Transformer) & 0.81 \\
Transformer-only & Tokenized input + Transformer & 0.91 \\
Hybrid backbone & 3D CNN stem + multiscale Transformer & 0.93 \\
\hline
Full ConvFormer3D-TAP & Hybrid + TAP modules & 0.96 \\
\hline
\end{tabularx}
\end{threeparttable}
\end{table}

Performance gains from TAP components reflect their complementary roles. Phase-aware sampling emphasizes high-motion transitions that help disambiguate adjacent long-axis and outflow-tract views. Masked spatiotemporal reconstruction regularizes learning by requiring recovery of missing anatomy and motion, improving robustness to common artifacts. Uncertainty-weighted multi-clip fusion down-weights ambiguous clips and stabilizes study-level predictions across the cardiac cycle. Together, these modules align training and inference with physiologically meaningful cine dynamics and reduce clinically typical error modes. Cine-specific design choices consistently outperform both CNN-only baselines and general-purpose video Transformers. The hybrid backbone, cine-aware sampling, reconstruction regularization, and uncertainty fusion address anatomical adjacency and acquisition variability, while remaining confusions are largely limited to clinically plausible pairs (e.g., CINE\_2 vs.\ CINE\_3; AV vs.\ LVOT). Future work includes self-supervised pretraining on larger cine corpora, explicit cardiac-phase modeling, and prospective multi-center evaluation, as well as end-to-end assessment of downstream impact on segmentation, biomarkers, and workflow efficiency.

Despite these strengths, several limitations warrant consideration. First, although the dataset is large and clinically diverse, all samples were drawn from a single health system, and broader multi-center evaluation will be necessary to fully characterize performance under wider vendor, protocol, and population variability. Second, while the model is robust to common motion artifacts and mild corruption, its performance under severe arrhythmia, mistriggering, or highly nonstandard acquisitions remains untested. Third, the current framework operates on 16-frame clips rather than full cycles, which—despite phase-aware sampling—may omit atypical temporal patterns in patients with complex physiology. Fourth, masked reconstruction improves representation learning but increases training complexity, and its optimal masking strategy may differ across cohorts. Finally, although study-level inference uses uncertainty-weighted fusion, clinical deployment would benefit from integrating explicit quality assessment, real-time feedback during acquisition, and prospective evaluation of downstream effects on segmentation, volumetry, and diagnostic biomarkers. Addressing these limitations represents an important direction for future research and translation.

Future work will focus on expanding evaluation to multi-center and multi-vendor cohorts, incorporating explicit cardiac-phase modeling to further disambiguate challenging views, and exploring large-scale self-supervised pretraining for improved domain generalization. Importantly, reliable cine view classification can serve as an initial step toward professional-grade, view-specific segmentation pipelines by enabling correct routing to specialized segmentation models and anatomical priors, thereby supporting more accurate downstream quantification for disease diagnosis. Integrating ConvFormer3D-TAP into full CMR analysis pipelines will enable prospective assessment of its impact on segmentation accuracy, volumetric and strain biomarkers, and workflow efficiency. Additional opportunities include real-time adaptive acquisition guidance, quality-aware clip selection, and joint modeling of view classification alongside segmentation, motion estimation, or chamber-level anatomical reasoning. Together, these directions aim to advance cine-aware representation learning toward robust, interpretable, and clinically integrated cardiac MRI automation.

\section{Conclusion}
\label{sec:conclusion}

Accurate cine cardiac MRI view recognition is a necessary first step for reliable automated CMR analysis, yet it remains difficult under real-world variability in acquisition protocols, artifacts, and plane prescription. We introduced ConvFormer3D-TAP, a cine-specific spatiotemporal framework that combines a 3D convolutional stem with multiscale attention and TAP modules (phase-aware sampling, masked reconstruction, and uncertainty-weighted multi-clip fusion) to improve robustness across the cardiac cycle. On 150{,}974 clinically acquired cine sequences spanning six standard views, the proposed method achieved 96\% validation accuracy with strong per-class performance and calibration, outperforming

\section*{Declarations}

\section*{Funding}
Funding was not received for this work.

\section*{Conflict of Interests}
The authors declare no conflicts of interests.

\section*{Consent for publication}
All authors consent to publication of the manuscript.

% \section*{Author Contributions}
% \textbf{Nafiseh~Ghaffar~Nia} led the study conception and design, developed the technical methodology and analytic workflow, performed the experiments and/or computational analyses, generated the figures and tables, and drafted the manuscript. \textbf{Adrienne~Kline} provided overall supervision and strategic direction, guided study design and interpretation, and critically revised the manuscript for intellectual content; she served as the corresponding author. \textbf{Vinesh~Appadurai} contributed cardiology domain expertise, supported interpretation of clinical relevance and implications. \textbf{Suchithra~V.} and \textbf{Chinmay~Rane} contributed engineering/implementation perspectives, and revised the manuscript. \textbf{Daniel~Pittman} and \textbf{James~Carr} supported data curation and quality control, assisted with experiments and/or validation, and contributed to manuscript editing. All authors reviewed the final manuscript and approved the submitted version.

%Bibliography
\FloatBarrier
\bibliographystyle{unsrt} 
\bibliography{references}  

\newpage

\appendix

\end{document}